\documentclass[manuscript]{acmart}

\AtBeginDocument{%
  }

\usepackage{subcaption}
\usepackage{wrapfig}

\setcopyright{none}
\copyrightyear{2024}
\acmYear{2024}
\title{Attention-Based Learning for Fluid State Interpolation and Editing in a Time-Continuous Framework}

\author{Bruno Roy}
\email{bruno.roy@autodesk.com}
\affiliation{%
  \institution{Autodesk Research}
  \city{Montreal}
  \country{Canada}
}

\begin{document}

\begin{teaserfigure}
    \centering
    \includegraphics[width=\linewidth]{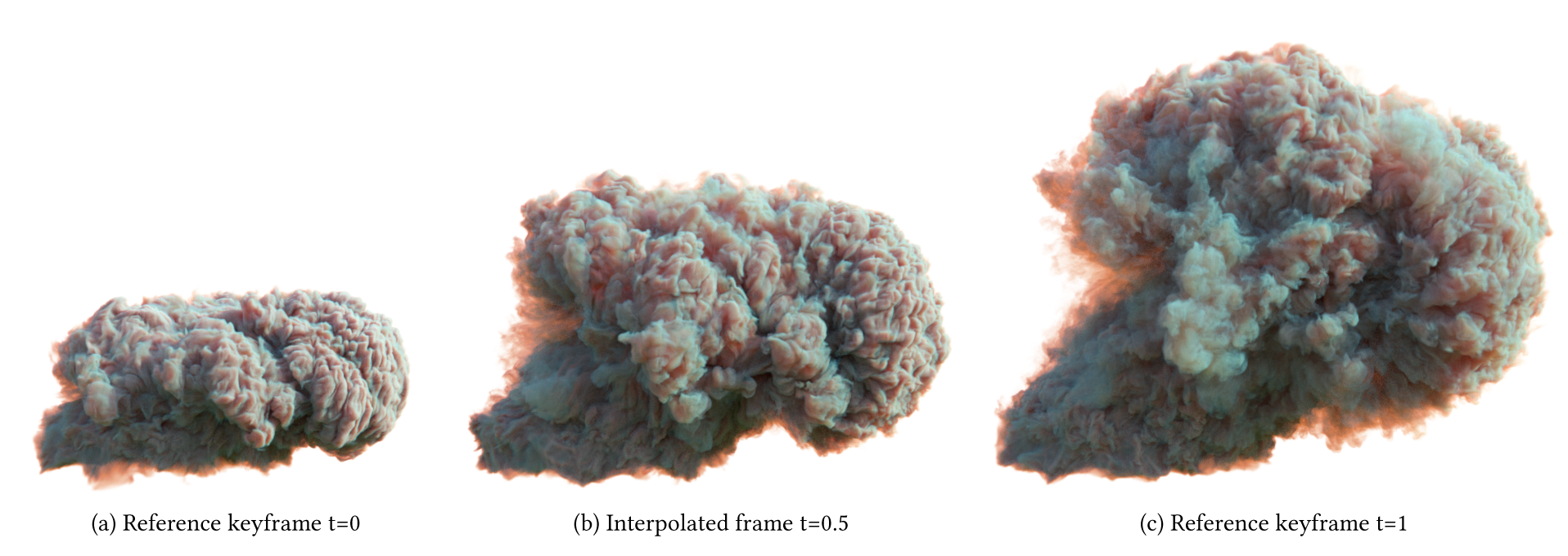}
    \caption{Given input keyframes, our approach interpolates substeps of a fluid simulation – resulting in a smooth and realistic animation.}
    \label{fig:teaser}
\end{teaserfigure}

\begin{abstract}
In this work, we introduce \textit{FluidsFormer}: a transformer-based approach for fluid interpolation within a continuous-time framework. By combining the capabilities of PITT and a residual neural network (RNN), we analytically predict the physical properties of the fluid state. This enables us to interpolate substep frames between simulated keyframes, enhancing the temporal smoothness and sharpness of animations. We demonstrate promising results for smoke interpolation and conduct initial experiments on liquids.
\end{abstract}

\maketitle

\section{Introduction}
As we advance into the era of generative AI, there has been a surge of interest in editing within the latent space, presenting a genuine challenge in providing controllable data-driven capabilities. Among many other areas in computer graphics, physics-based animation remains particularly challenging to edit as it relies on complex physics rules and principles that need to be satisfied for realism. Another challenge posed by these physics-based phenomena, particularly within the visual effects industry, is the need to adapt these principles to align with artistic direction, as observed in animation films. Striking a balance between realism and controllability poses difficulty in providing tools for editing natural phenomena such as fluids.
For several decades, numerous researchers have endeavored to enhance the controllability and flexibility of fluid editing – transitioning from local editing of keyframes~\cite{pan2013interactive} to flow-based methods~\cite{sato2018editing}. While the latter showed promise in terms of controllability, some explored optical flow-based approaches to interpolate Eulerian~\cite{thuerey2016interpolations} and particle-based fluids~\cite{roy2021neural} as novel means of creating and editing such natural phenomena. Although also promising, these flow-based approaches remained highly dependent on numerical solvers, rendering them still fairly computationally expensive.
More recently, data-driven methods have emerged to simulate and control fluids at a reduced cost. Introduced to computer graphics approximately a decade ago, \cite{ladicky2015data} proposed a novel approach to computing particle acceleration using a regression forest. Subsequently, a significant advancement was made by utilizing LSTM-based methods~\cite{wiewel2019latent} to handle and compute pressure changes as sequential data. Similarly, other works were introduced to address the pressure projection step using CNNs~\cite{yang2016data, tompson2017accelerating}. Techniques were also proposed to synthesize smoke from pre-computed patches~\cite{chu2017data}, generate super-resolution flows using GANs~\cite{xie2018tempogan}, and enhance diffusion behavior and liquid splashes~\cite{um2018liquid}. In recent years, methods have been introduced to improve the apparent resolution of smoke~\cite{bai2020dynamic} and particle-based liquids~\cite{roy2021neural}. Although our primary objective remains to enhance the controllability of fluid editing, our approach shares similarities with that of \cite{thuerey2016interpolations} and \cite{roy2021neural} as we aim to interpolate fluids in a data-driven manner using the advection scheme of Eulerian simulations.

\begin{figure}[t]
    \centering
    \includegraphics[width=\linewidth]{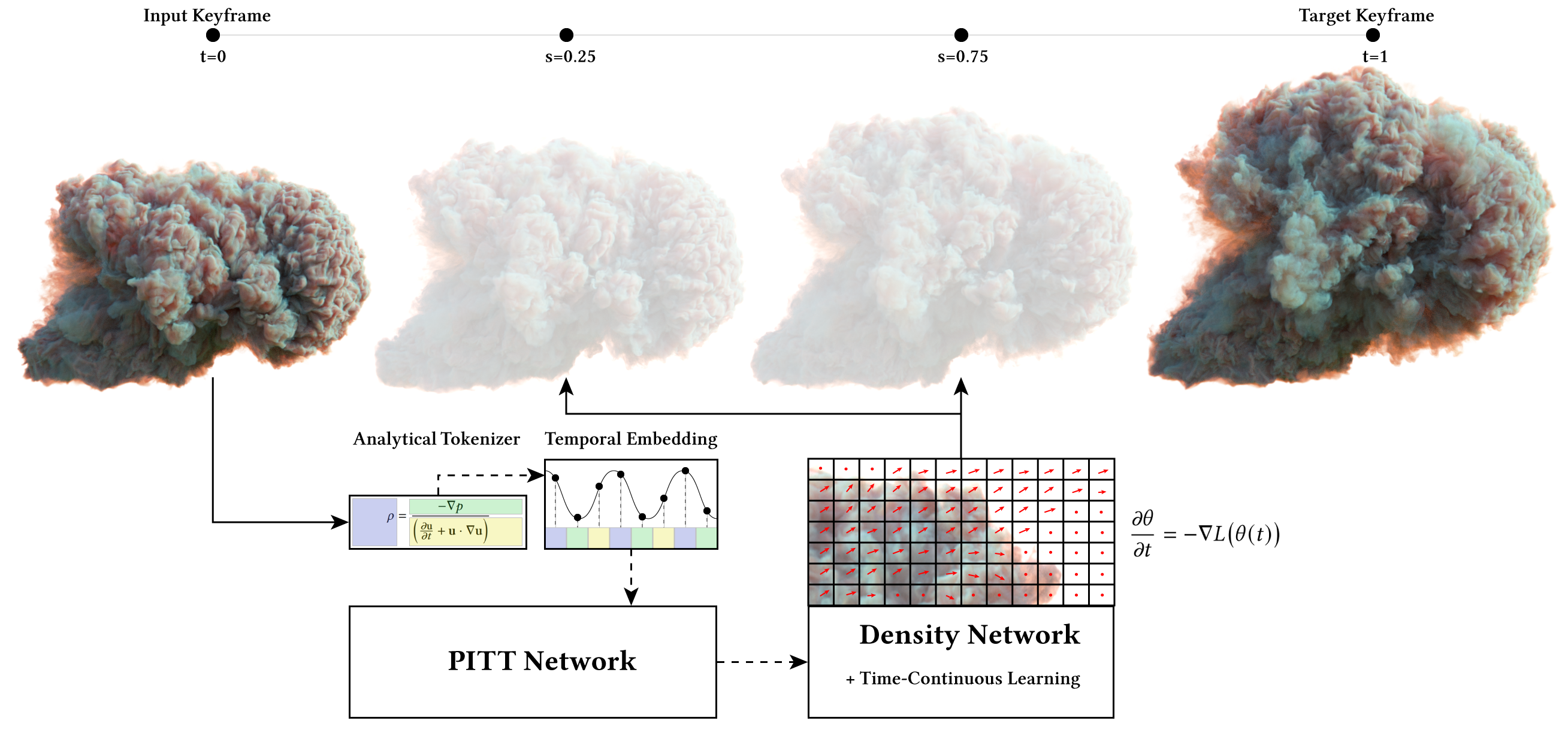}
    \caption{Here is an overview of the \textit{FluidsFormer} approach. A conventional Eulerian solver is used to produce keyframes, along with their respective volumetric properties, using a large timestep. We leverage most of the PITT network to tokenize and embed the partial differential equations (PDEs) needed to compute the density at any given time between the reference frames. Finally, our density network predicts the advected densities at substeps (e.g., $s=[0.25,0.75]$).}
    \label{fig:overview}
\end{figure}

\section{Our Method}
Although Transformer-based networks have been primarily introduced for natural language processing (NLP) and text generation, they offer interesting properties for sequential data in general. In our method, we propose to leverage an attention-based architecture within a continuous-time framework to learn and interpolate simulation properties per frame in an approximate analytical manner. In the following sections, we will outline the details spanning from data preparation to network architecture and training using a transformer-based encoder-decoder network. We will also highlight a few concrete use cases to introduce novel ways of generating and editing fluids.

\subsection{Data Preparation}\label{seq:data-prep}
The data preparation is divided into two main steps: (1) generating the temporal embeddings for the fluid element’s states and (2) handling the tokenization process in a physics-adapted context.
\paragraph{\textbf{Physics-Adapted Tokenization}}
The tokenization process is performed by parsing and splitting the Navier-Stokes equation into components (see Eq.~\ref{eq:navier-stokes}) – we intentionally omit the viscosity term to simplify the related operations.
\begin{equation}
\label{eq:navier-stokes}
\rho \left( \frac{\partial \mathbf{u}}{\partial t} + \mathbf{u} \cdot \nabla \mathbf{u} \right) = - \nabla p
\end{equation}
% \begin{equation}
% \label{eq:navier-stokes}
% \rho = \frac{-\nabla p}{\left( \frac{\partial \mathbf{u}}{\partial t} + \mathbf{u} \cdot \nabla \mathbf{u} \right)}
% \end{equation}
% \begin{equation}
% \frac{\partial\mathbf{\theta}}{\partial t}=-\nabla L\bigl(\mathbf{\theta}(t)\bigr)
% \end{equation}
\paragraph{\textbf{Temporal Embedding}}
We learn the latent embedding of the advection part of the governing equation using standard multi-head self-attention blocks from the PITT architecture~\cite{lorsung2024physics}. That way, our model is capable of learning to interpolate physics properties analytically (e.g., densities $\rho$). As we do not consider the equilibrium equation (i.e., $\nabla \cdot \mathbf{u} = 0$), we handle the volume preservation part by simply penalizing in our loss function the solutions diverging too much from the reference. 
\subsection{Network Architecture}
Our network architecture is composed of two stacked networks: (1) the pre-trained PITT network (transformer-based network) for solving the governing partial differential equation of the fluid dynamics and (2) the density network (RNN) to learn and predict the time-continuous density for the substeps between the input keyframes. We use the 18-layer architecture for our residual neural network as it gives us a decent performance (i.e., training/inference speed and accuracy) while reducing the requirement for an enormous dataset. Same as the ResNet original paper~\cite{he2016deep}, our network is composed of one 7x7 convolution layer (including a 3x3 max pooling layer), 8 pairs of 3x3 convolution layers with respectively 64, 128, 256, and 512 kernels, and one last fully connected layer using a softmax activation function for average pooling (the rest of layers are using a ReLU activation function).
\subsection{Training and Dataset}
The density network is trained on normalized data $[-1, 1]$ outputted from the PITT network. During training, each simulation scenario is processed through the PITT network to generate the latent embedding of the non-viscous governing equation and provides an analytically-driven approximation to our density network to predict the correct density in space and time. To update the parameters, we use a Huber loss function $L_\delta$ minimizing a single term considering the possible outliers (with some regularization term): the difference in density between the ground truth $\rho$ and the prediction $\hat{\rho}$.
\begin{equation}
L_\delta\bigl(\rho,\hat{\rho}\bigr)=
\begin{cases}
\frac{1}{2}(\rho-\hat{\rho})^2 & \bigl|\rho-\hat{\rho}\bigr|\leq\delta,\\
\delta\bigl|\rho-\hat{\rho}\bigr|-\frac{1}{2}\delta^2 & \text{otherwise}.
\end{cases}
\end{equation}
As the reference density is advected using the divergence-free velocity field, we already the volume preservation law.

We validated and tweaked the hyperparameters of our model using a 2D dataset of laminar and turbulent flows generated with OpenFOAM~\cite{jasak2009openfoam}. Then we generated volumetric data from Eulerian simulations using Bifrost. The volumetric data points are constituted of a position (center of the cell), a velocity, and a density. The volumetric dataset is composed of 1000 (800 for training, 100 for validation, and 100 for testing) smoke inflow and emission simulations of 50 frames in length, without and with a single obstacle placed at random locations. As for most RNN architectures, we need a significant amount of data to properly generalize without dissipating the small-scale details in the simulation (e.g., second-order vorticity).

\subsection{Continuous-Time Learning}
Similarly to \cite{chen2024contiformer}, our architecture uses a continuous-time multi-head attention module to transform time-varying sequence relationships into vectors of queries Q, keys K, and values V. The purpose of this module is to output a continuous dynamic flow evolving throughout the data points. However, as opposed to \cite{chen2024contiformer} and inspired by \cite{deleu2022continuous}, we formulate the learning algorithm to follow the input velocity during training – allowing our model to converge faster and to reflect the analytical framework as discussed in Sec.~\ref{seq:data-prep}. The updated gradients are then used to update the parameters defining the fluid’s behavior. In other words, we train our model to evaluate the density based on the advection term and the velocities of the governing equation of the input system. An inherent advantage of analytically learning density advection is the flexibility to dynamically choose the discretization during the inference stage as required. Essentially, by employing the pre-trained PITT network, we evenly divide the time interval between two keyframes into a specified number of substeps $S$. Subsequently, we evaluate the density at these time points while considering the initial conditions. The density $\rho$ is computed at location $x$ by advecting the previous density with respect to the input velocity.

\section{Various Applications}
\paragraph{\textbf{Eulerian Fluids Interpolation}}
Our main goal with this approach is to propose a continuous-time transformer model capable of learning the underlying dynamics of fluid systems for interpolation purposes. The idea behind interpolating fluids is to generate a visually appealing and temporally smooth animation using only a few keyframes at large timesteps. Our interpolation methods will fill in between similarly to simulate the substeps between the provided reference keyframes (as shown in Fig.~\ref{fig:teaser}).
\paragraph{\textbf{Generating using Variants}}
In this last use case, we take advantage of the generated tree structure to combine keyframes using Boolean operations such as addition, subtraction, and intersection. Using the volumetric data (i.e., properly stored in grid cells), we can mix multiple keyframes into a single target in our approach and produce a completely new animation.

\paragraph{\textbf{Tree-Based Variants}}
\begin{wrapfigure}[12]{r}{0.5\textwidth}
    \centering
    \includegraphics[width=0.45\textwidth]{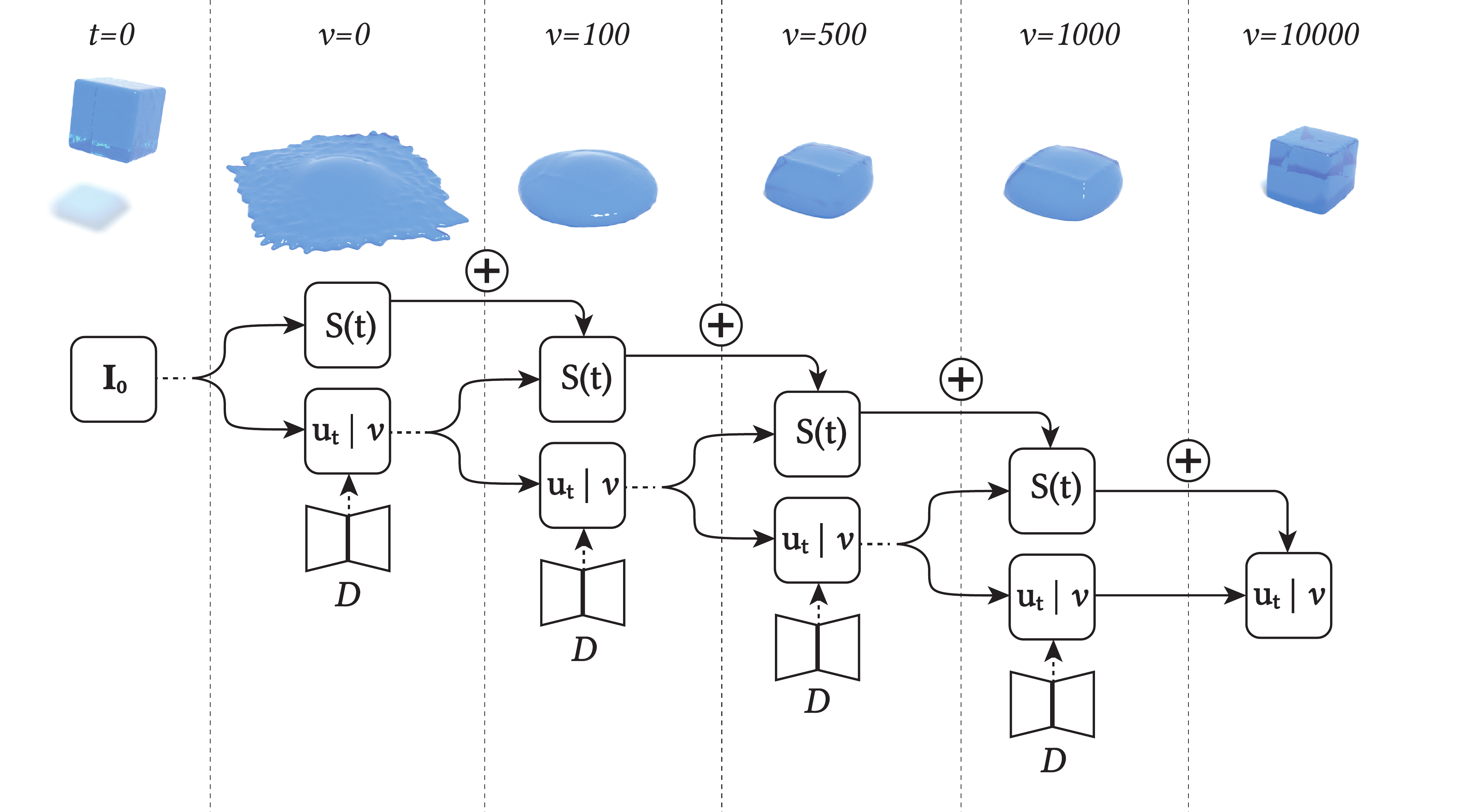}
    \caption{Combining \textit{FluidsFormer} with an explicit solver to generate viscosity variations for liquids.}
    \label{fig:applications}
\end{wrapfigure}
For this use case, for each frame, we output multiple probable solutions using top-k sampling along with diverse Beam Search on the decoder side to encourage diversity in the generated sequences. From this set of solutions, we build a tree structure that allows us to branch out at any node to produce variants of a single simulation while preserving the initial conditions.
We also performed a few early experiments combining our approach to an explicit solver for liquid simulations. As opposed to other presented use cases, we tested our approach to interpolate between various viscosity states. As shown in Fig.~\ref{fig:applications}, we use our approach to produce five (5) variations of the same simulation but using different viscosity values $\nu$ (0 being the less viscous and 10000 the most). Starting with the same initial conditions $\textbf{I}_0$, we branch a new variation of the current state of the velocity by predicting the next sequence of velocities according to a certain viscosity threshold (e.g., $\nu\in\{0, 100, 500, 1000, 10000\}$). To learn and generate these sequences based on the current state, we have trained a viscosity network $D$ (i.e., including the viscosity term in the PITT embedding as input to a residual neural network) to match similarities between fluid characteristics (e.g., viscosity) and their corresponding velocities. Between each reference keyframe generated by the explicit solver, our network $D$ interpolates viscosities to generate the substeps which are guided by the computed velocity field.

\section{Conclusion}
While our approach still relies on a coarse numerical simulation, we are confident that it introduces novel and less-linear ways of interacting with fluids. In future work, we aim to explore the accuracy of employing Transformer-based networks like PITT to replace conventional numerical solvers for simulating natural phenomena in visual effects.

\bibliographystyle{ACM-Reference-Format}
\bibliography{paper_arxiv}


\begin{thebibliography}{17}

%%% the Web address.

\ifx \showCODEN    \undefined \def \showCODEN     #1{\unskip}     \fi
\ifx \showDOI      \undefined \def \showDOI       #1{#1}\fi
\ifx \showISBNx    \undefined \def \showISBNx     #1{\unskip}     \fi
\ifx \showISBNxiii \undefined \def \showISBNxiii  #1{\unskip}     \fi
\ifx \showISSN     \undefined \def \showISSN      #1{\unskip}     \fi
\ifx \showLCCN     \undefined \def \showLCCN      #1{\unskip}     \fi
\ifx \shownote     \undefined \def \shownote      #1{#1}          \fi
\ifx \showarticletitle \undefined \def \showarticletitle #1{#1}   \fi
\ifx \showURL      \undefined \def \showURL       {\relax}        \fi
% invisible to TeX
\providecommand\bibfield[2]{#2}
\providecommand\bibinfo[2]{#2}
\providecommand\natexlab[1]{#1}
\providecommand\showeprint[2][]{arXiv:#2}

\bibitem[Bai et~al\mbox{.}(2020)]%
        {bai2020dynamic}
\bibfield{author}{\bibinfo{person}{Kai Bai}, \bibinfo{person}{Wei Li}, \bibinfo{person}{Mathieu Desbrun}, {and} \bibinfo{person}{Xiaopei Liu}.} \bibinfo{year}{2020}\natexlab{}.
\newblock \showarticletitle{Dynamic upsampling of smoke through dictionary-based learning}.
\newblock \bibinfo{journal}{\emph{ACM Transactions on Graphics (TOG)}} \bibinfo{volume}{40}, \bibinfo{number}{1} (\bibinfo{year}{2020}), \bibinfo{pages}{1--19}.
\newblock


\bibitem[Chen et~al\mbox{.}(2024)]%
        {chen2024contiformer}
\bibfield{author}{\bibinfo{person}{Yuqi Chen}, \bibinfo{person}{Kan Ren}, \bibinfo{person}{Yansen Wang}, \bibinfo{person}{Yuchen Fang}, \bibinfo{person}{Weiwei Sun}, {and} \bibinfo{person}{Dongsheng Li}.} \bibinfo{year}{2024}\natexlab{}.
\newblock \showarticletitle{ContiFormer: Continuous-time transformer for irregular time series modeling}.
\newblock \bibinfo{journal}{\emph{Advances in Neural Information Processing Systems}}  \bibinfo{volume}{36} (\bibinfo{year}{2024}).
\newblock


\bibitem[Chu and Thuerey(2017)]%
        {chu2017data}
\bibfield{author}{\bibinfo{person}{Mengyu Chu} {and} \bibinfo{person}{Nils Thuerey}.} \bibinfo{year}{2017}\natexlab{}.
\newblock \showarticletitle{Data-driven synthesis of smoke flows with CNN-based feature descriptors}.
\newblock \bibinfo{journal}{\emph{ACM Transactions on Graphics (TOG)}} \bibinfo{volume}{36}, \bibinfo{number}{4} (\bibinfo{year}{2017}), \bibinfo{pages}{1--14}.
\newblock


\bibitem[Deleu et~al\mbox{.}(2022)]%
        {deleu2022continuous}
\bibfield{author}{\bibinfo{person}{Tristan Deleu}, \bibinfo{person}{David Kanaa}, \bibinfo{person}{Leo Feng}, \bibinfo{person}{Giancarlo Kerg}, \bibinfo{person}{Yoshua Bengio}, \bibinfo{person}{Guillaume Lajoie}, {and} \bibinfo{person}{Pierre-Luc Bacon}.} \bibinfo{year}{2022}\natexlab{}.
\newblock \showarticletitle{Continuous-time meta-learning with forward mode differentiation}.
\newblock \bibinfo{journal}{\emph{arXiv preprint arXiv:2203.01443}} (\bibinfo{year}{2022}).
\newblock


\bibitem[He et~al\mbox{.}(2016)]%
        {he2016deep}
\bibfield{author}{\bibinfo{person}{Kaiming He}, \bibinfo{person}{Xiangyu Zhang}, \bibinfo{person}{Shaoqing Ren}, {and} \bibinfo{person}{Jian Sun}.} \bibinfo{year}{2016}\natexlab{}.
\newblock \showarticletitle{Deep residual learning for image recognition}. In \bibinfo{booktitle}{\emph{Proceedings of the IEEE conference on computer vision and pattern recognition}}. \bibinfo{pages}{770--778}.
\newblock


\bibitem[Jasak(2009)]%
        {jasak2009openfoam}
\bibfield{author}{\bibinfo{person}{Hrvoje Jasak}.} \bibinfo{year}{2009}\natexlab{}.
\newblock \showarticletitle{OpenFOAM: Open source CFD in research and industry}.
\newblock \bibinfo{journal}{\emph{International Journal of Naval Architecture and Ocean Engineering}} \bibinfo{volume}{1}, \bibinfo{number}{2} (\bibinfo{year}{2009}), \bibinfo{pages}{89--94}.
\newblock


\bibitem[Ladick{\`y} et~al\mbox{.}(2015)]%
        {ladicky2015data}
\bibfield{author}{\bibinfo{person}{L'ubor Ladick{\`y}}, \bibinfo{person}{SoHyeon Jeong}, \bibinfo{person}{Barbara Solenthaler}, \bibinfo{person}{Marc Pollefeys}, {and} \bibinfo{person}{Markus Gross}.} \bibinfo{year}{2015}\natexlab{}.
\newblock \showarticletitle{Data-driven fluid simulations using regression forests}.
\newblock \bibinfo{journal}{\emph{ACM Transactions on Graphics (TOG)}} \bibinfo{volume}{34}, \bibinfo{number}{6} (\bibinfo{year}{2015}), \bibinfo{pages}{1--9}.
\newblock


\bibitem[Lorsung et~al\mbox{.}(2024)]%
        {lorsung2024physics}
\bibfield{author}{\bibinfo{person}{Cooper Lorsung}, \bibinfo{person}{Zijie Li}, {and} \bibinfo{person}{Amir Barati~Farimani}.} \bibinfo{year}{2024}\natexlab{}.
\newblock \showarticletitle{Physics Informed Token Transformer for Solving Partial Differential Equations}.
\newblock \bibinfo{journal}{\emph{Machine Learning: Science and Technology}} (\bibinfo{year}{2024}).
\newblock


\bibitem[Pan et~al\mbox{.}(2013)]%
        {pan2013interactive}
\bibfield{author}{\bibinfo{person}{Zherong Pan}, \bibinfo{person}{Jin Huang}, \bibinfo{person}{Yiying Tong}, \bibinfo{person}{Changxi Zheng}, {and} \bibinfo{person}{Hujun Bao}.} \bibinfo{year}{2013}\natexlab{}.
\newblock \showarticletitle{Interactive localized liquid motion editing}.
\newblock \bibinfo{journal}{\emph{ACM Transactions on Graphics (TOG)}} \bibinfo{volume}{32}, \bibinfo{number}{6} (\bibinfo{year}{2013}), \bibinfo{pages}{1--10}.
\newblock


\bibitem[Roy et~al\mbox{.}(2021)]%
        {roy2021neural}
\bibfield{author}{\bibinfo{person}{Bruno Roy}, \bibinfo{person}{Pierre Poulin}, {and} \bibinfo{person}{Eric Paquette}.} \bibinfo{year}{2021}\natexlab{}.
\newblock \showarticletitle{Neural upflow: A scene flow learning approach to increase the apparent resolution of particle-based liquids}.
\newblock \bibinfo{journal}{\emph{Proceedings of the ACM on Computer Graphics and Interactive Techniques}} \bibinfo{volume}{4}, \bibinfo{number}{3} (\bibinfo{year}{2021}), \bibinfo{pages}{1--26}.
\newblock


\bibitem[Sato et~al\mbox{.}(2018)]%
        {sato2018editing}
\bibfield{author}{\bibinfo{person}{Syuhei Sato}, \bibinfo{person}{Yoshinori Dobashi}, {and} \bibinfo{person}{Tomoyuki Nishita}.} \bibinfo{year}{2018}\natexlab{}.
\newblock \showarticletitle{Editing fluid animation using flow interpolation}.
\newblock \bibinfo{journal}{\emph{ACM Transactions on Graphics (TOG)}} \bibinfo{volume}{37}, \bibinfo{number}{5} (\bibinfo{year}{2018}), \bibinfo{pages}{1--12}.
\newblock


\bibitem[Thuerey(2016)]%
        {thuerey2016interpolations}
\bibfield{author}{\bibinfo{person}{Nils Thuerey}.} \bibinfo{year}{2016}\natexlab{}.
\newblock \showarticletitle{Interpolations of smoke and liquid simulations}.
\newblock \bibinfo{journal}{\emph{ACM Transactions on Graphics (TOG)}} \bibinfo{volume}{36}, \bibinfo{number}{1} (\bibinfo{year}{2016}), \bibinfo{pages}{1--16}.
\newblock


\bibitem[Tompson et~al\mbox{.}(2017)]%
        {tompson2017accelerating}
\bibfield{author}{\bibinfo{person}{Jonathan Tompson}, \bibinfo{person}{Kristofer Schlachter}, \bibinfo{person}{Pablo Sprechmann}, {and} \bibinfo{person}{Ken Perlin}.} \bibinfo{year}{2017}\natexlab{}.
\newblock \showarticletitle{Accelerating eulerian fluid simulation with convolutional networks}. In \bibinfo{booktitle}{\emph{International Conference on Machine Learning}}. PMLR, \bibinfo{pages}{3424--3433}.
\newblock


\bibitem[Um et~al\mbox{.}(2018)]%
        {um2018liquid}
\bibfield{author}{\bibinfo{person}{Kiwon Um}, \bibinfo{person}{Xiangyu Hu}, {and} \bibinfo{person}{Nils Thuerey}.} \bibinfo{year}{2018}\natexlab{}.
\newblock \showarticletitle{Liquid splash modeling with neural networks}. In \bibinfo{booktitle}{\emph{Computer Graphics Forum}}, Vol.~\bibinfo{volume}{37}. Wiley Online Library, \bibinfo{pages}{171--182}.
\newblock


\bibitem[Wiewel et~al\mbox{.}(2019)]%
        {wiewel2019latent}
\bibfield{author}{\bibinfo{person}{Steffen Wiewel}, \bibinfo{person}{Moritz Becher}, {and} \bibinfo{person}{Nils Thuerey}.} \bibinfo{year}{2019}\natexlab{}.
\newblock \showarticletitle{Latent space physics: Towards learning the temporal evolution of fluid flow}. In \bibinfo{booktitle}{\emph{Computer graphics forum}}, Vol.~\bibinfo{volume}{38}. Wiley Online Library, \bibinfo{pages}{71--82}.
\newblock


\bibitem[Xie et~al\mbox{.}(2018)]%
        {xie2018tempogan}
\bibfield{author}{\bibinfo{person}{You Xie}, \bibinfo{person}{Erik Franz}, \bibinfo{person}{Mengyu Chu}, {and} \bibinfo{person}{Nils Thuerey}.} \bibinfo{year}{2018}\natexlab{}.
\newblock \showarticletitle{tempoGAN: A temporally coherent, volumetric GAN for super-resolution fluid flow}.
\newblock \bibinfo{journal}{\emph{ACM Transactions on Graphics (TOG)}} \bibinfo{volume}{37}, \bibinfo{number}{4} (\bibinfo{year}{2018}), \bibinfo{pages}{1--15}.
\newblock


\bibitem[Yang et~al\mbox{.}(2016)]%
        {yang2016data}
\bibfield{author}{\bibinfo{person}{Cheng Yang}, \bibinfo{person}{Xubo Yang}, {and} \bibinfo{person}{Xiangyun Xiao}.} \bibinfo{year}{2016}\natexlab{}.
\newblock \showarticletitle{Data-driven projection method in fluid simulation}.
\newblock \bibinfo{journal}{\emph{Computer Animation and Virtual Worlds}} \bibinfo{volume}{27}, \bibinfo{number}{3-4} (\bibinfo{year}{2016}), \bibinfo{pages}{415--424}.
\newblock


\end{thebibliography}

\end{document}